\documentclass[letterpaper, 10 pt, conference]{ieeeconf}  
\usepackage{amsmath,amssymb,amsfonts}
\usepackage{url}
\usepackage[algo2e,linesnumbered,ruled,vlined]{algorithm2e}
\usepackage[dvipsnames]{xcolor}
\usepackage{cleveref}
\usepackage{booktabs}
\usepackage{multirow}
\usepackage{subcaption}
\usepackage{bm}
\usepackage{gensymb}

\IEEEoverridecommandlockouts                              

\usepackage{graphics} 
\usepackage{epsfig} 

\title{\LARGE \bf
    SCRAMPPI: Contingency-Constrained Planning using Hamilton-Jacobi Reachability
}

\author{Raj Harshit Srirangam, Leonard Jung, Rohith Poola, and Michael Everett
\thanks{All authors are with Northeastern University, Boston, MA 02115, USA 
        {\tt\small \{srirangam.r, jung.le, poola.r, m.everett\}@northeastern.edu}}%
}

\begin{document}
\newcommand{\manci}{{m_{\text{anci}}}}
\newcommand{\Manci}{{M_{\text{anci}}}}
\newcommand{\m}{m}
\newcommand{\M}{M}
\newcommand{\unomzero}{\mathbf{u}'_{0}}
\newcommand{\scenarioOne}{Limited Sensing Environment}
\newcommand{\scenarioTwo}{Ice Slip Environment}
\newcommand{\TaskSuccess}{Success Rate}
\newcommand{\TTGoal}{Goal Time}
\newcommand{\SafeRate}{Safety Violations/Episode}
\newcommand{\PlanningLat}{Average Planning Time}
\newcommand{\TODO}[1]{{\color{red}\textbf{TODO:} #1}}

\definecolor{darkblue}{rgb}{0.0, 0.0, 0.55}
\maketitle
\thispagestyle{empty}
\pagestyle{empty}

\begin{abstract}

    Autonomous robots commonly aim to achieve a nominal objective while minimizing a cost. Without contingency preparation, this can leave them vulnerable to mission failure. This is formalized as a trajectory optimization problem over the nominal cost with a contingency constraint: from each checked state along the nominal plan, an admissible contingency maneuver must exist that reaches a designated recovery target within a prescribed contingency horizon. Existing methods either optimize contingency trajectories alongside the nominal plan or evaluate this constraint through computationally expensive nested sampling-based searches. Instead, we introduce SCRAMPPI, which represents this requirement as a reach-avoid problem and leverages Hamilton--Jacobi (HJ) reachability analysis to evaluate contingency feasibility. By either computing a reach-avoid value function offline or updating it only as the environment is revealed, and integrating it with model predictive path integral (MPPI) control via resampling-based rollouts, SCRAMPPI provides contingency-feasibility evaluations while still planning in 15-20 hz. Finally, we present simulated and hardware experiments demonstrating real-time nominal planning and execution of contingency maneuvers in mobile-robot navigation, alongside a five-dimensional friction-limited recovery simulation.
\end{abstract}

%
\section{Introduction}
\label{sec:introduction}
Autonomous mobile robots deployed in inspection, search-and-rescue, or surveillance tasks routinely face situations where the current mission must be abandoned. A sensor fault, a low-battery warning, or an operator abort command may require the robot to retreat to a designated recovery location such as a charging station, a landing zone, or a communication relay point. Unlike emergency stopping or local obstacle avoidance, this contingency maneuver requires the robot to actively navigate to a specific set of states while avoiding obstacles. Crucially, it must remain feasible along the nominal trajectory: if the robot enters a state from which no admissible maneuver can reach the recovery target within a bounded time horizon, the contingency requirement can no longer be met.

This type of contingency-constrained planning is difficult for several reasons. First, the contingency requirement is not given by a simple geometric state constraint like obstacle avoidance; it requires that a dynamically feasible maneuver to a target set exists from each candidate state. Checking this by planning a contingency from every state along the nominal plan introduces nested searches within an already constrained trajectory optimization. Second, the nominal planner cannot operate freely in the obstacle-free space; it is restricted to the subset of states from which a contingency is possible. This subset is not defined by simple geometric boundaries: it depends on the robot's dynamics, control authority, obstacle geometry, and time horizon. Third, in unknown environments with a limited sensing horizon, this feasible set evolves as obstacles and recovery targets are revealed. 

Existing contingency planning methods address hazard avoidance, stopping, or collision-free braking \cite{alsterda2019contingency,tordesillas2022faster,pek2021failsafe}, while branching methods plan over possible agent behaviors \cite{chen2022branch}. Contingency-MPPI \cite{jung2025contingency} explicitly constrains nominal plans to maintain access to a designated recovery target, but checks this requirement through nested sampling. This inherits a fundamental asymmetry: finding an admissible contingency demonstrates feasibility under the checked model, but failing to find one does not establish infeasibility. Increasing the sampling budget reduces search failures at additional computational cost. Hamilton--Jacobi (HJ) reachability has also been combined with sampling-based control for obstacle avoidance \cite{borquez2025dualguard} and used for navigation in unknown environments \cite{bajcsy2019efficient}. Here, we use its reach-avoid formulation to evaluate Contingency-MPPI's trajectory-wise recovery requirement.

\begin{figure}[t]
    \centering
    \includegraphics[width=0.98\columnwidth]{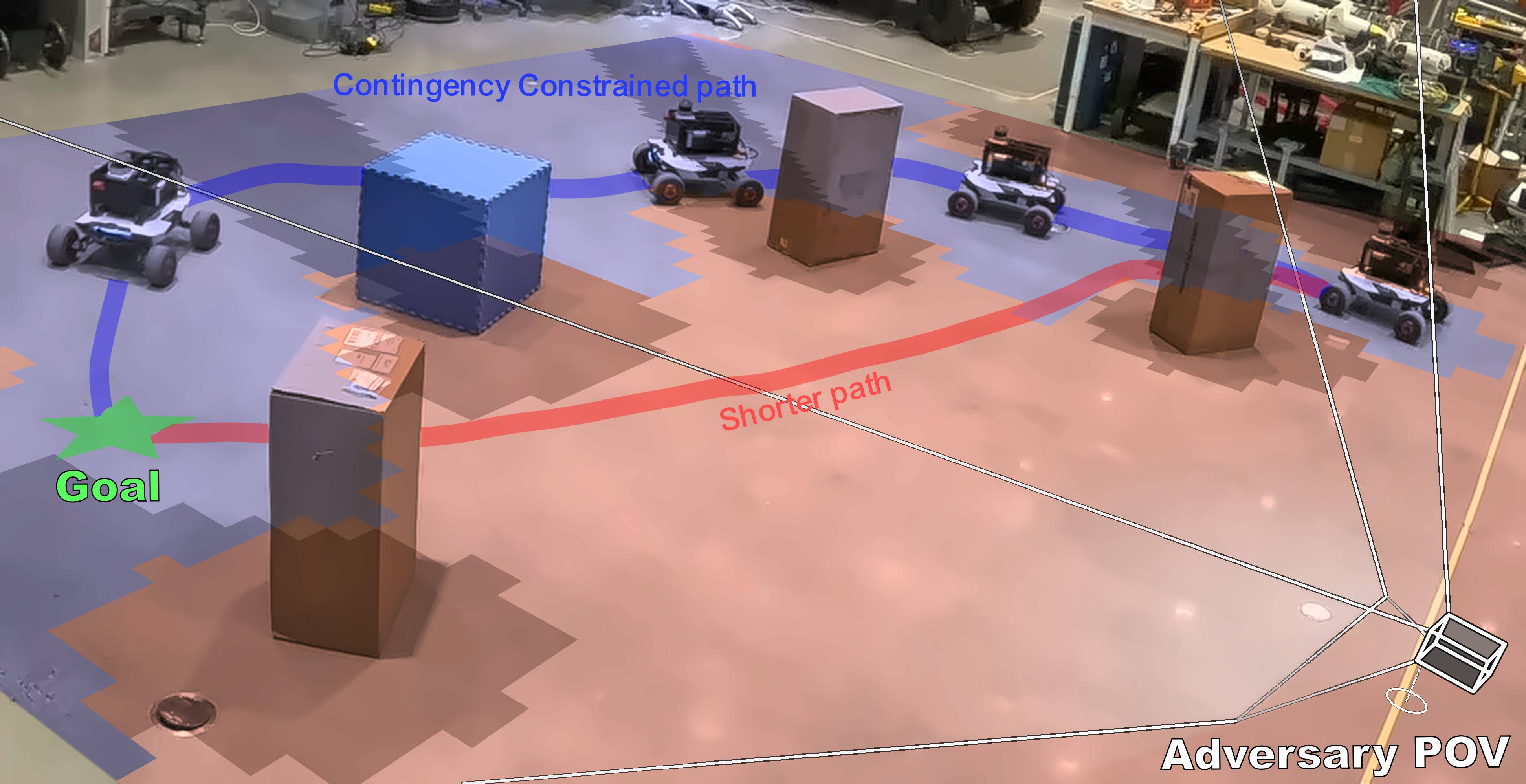}
    \caption{Contingency-constrained navigation for an occlusion based retreat task. The robot plans toward its nominal goal while retaining a retreat to a \textcolor{darkblue}{recovery target} occluded from an adversary. The \textcolor{blue}{feasible set} represents states from which this target can be reached within the contingency horizon.
    }
    \label{fig:overview}
\end{figure}

\begin{figure*}[t!]
    \centering
    \includegraphics[width=0.95\linewidth]{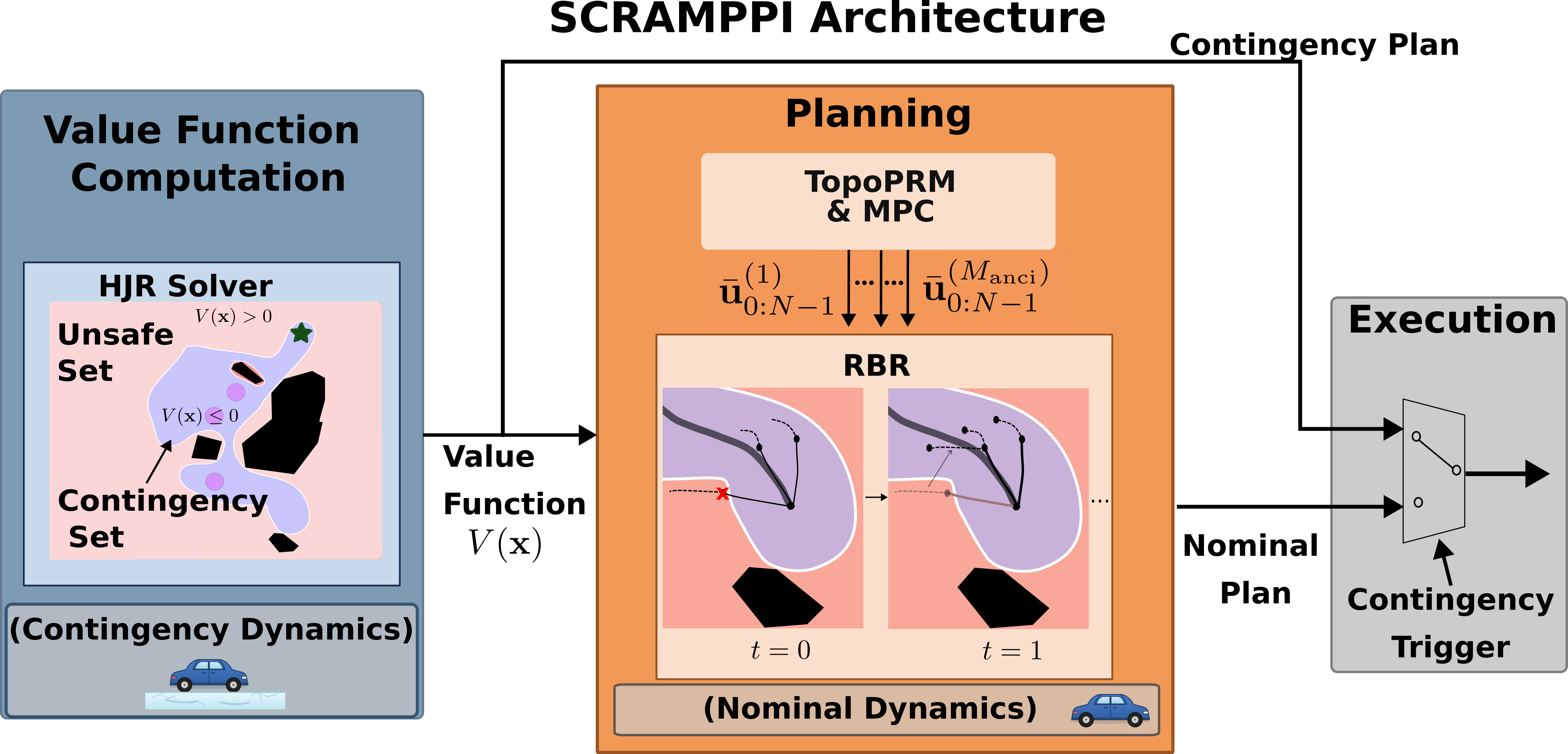}
    \caption{\textbf{SCRAMPPI Architecture}: (From left to right) We compute a finite-horizon reach-avoid value function from the recovery target, avoid set, and contingency dynamics. Together with the nominal goal, these are passed to our planner, which uses NMPC to seed a multimodal MPPI optimizer with resampling-based rollouts. The robot executes a validated nominal command or, upon a task-defined trigger, switches to HJ feedback to reach the recovery target within the remaining contingency horizon.}
    \label{fig:process}
\end{figure*}

The key insight is that the task-defined recovery requirement is a reach-avoid problem. Under the modeled dynamics and reach-avoid assumptions, the zero sub-level set of the HJ value function characterizes states from which the target is reachable within the contingency horizon while avoiding forbidden states \cite{bansal2017hamilton}. This replaces a sampled recovery search from every nominal state with an HJ value-function query; the gradient also supplies contingency feedback. Grid size grows exponentially with state dimension, limiting the method to problems within available memory and solve-time budgets. Within these limits, GPU-based solvers and batched queries enable real-time nominal planning, with HJ solves performed before operation or when the recovery problem changes.
We present Safe Contingency Reach-Avoid MPPI (SCRAMPPI), which integrates numerical HJ value-function queries into MPPI while retaining its nominal objective. Group-local resampling reuses surviving rollout prefixes \cite{yin2025safe}, and the selected command is checked before execution. A task-defined trigger activates HJ recovery feedback. For navigation with map revelation, the value function is updated over a conservative occupancy map. Figure~\ref{fig:process} outlines the framework.

Our contributions are as follows:
\begin{enumerate}
\item A contingency-constrained MPPI framework using finite-horizon HJ queries to evaluate task-defined recovery along nominal plans.
\item GPU-based reachability and batched queries integrated with multimodal resampling for real-time nominal planning within tractable grid sizes.
\item Baseline comparisons in navigation and five-dimensional friction-loss recovery, and an onboard mobile-robot occlusion-retreat demonstration.
\end{enumerate}

\section{Related Work}
\label{sec:related_work}

\subsection{Contingency Planning}

Contingency planning methods maintain backup maneuvers alongside a nominal plan so that the robot can respond to sudden changes. The type of guarantee varies across methods.

Several approaches address collision avoidance or emergency stopping. Contingency-MPC \cite{alsterda2019contingency,alsterda2021contingency} jointly optimizes nominal and contingency trajectories coupled at the first control action, accounting for hazards such as pop-out obstacles or loss of traction. FASTER \cite{tordesillas2022faster} maintains a committed backup trajectory in known-free space that terminates at rest. These formulations differ from requiring a maneuver to a designated recovery target from every nominal state.

A second class of methods handles multi-agent interaction. Branch MPC \cite{chen2022branch} and related approaches \cite{li2023marc} construct scenario trees over possible agent behaviors and optimize risk-aware plans. Their emphasis is interaction under behavioral uncertainty rather than a prescribed recovery destination.

Contingency-MPPI \cite{jung2025contingency} is the most closely related work. It nests a contingency MPPI planner inside a nominal MPPI planner: from every state along a candidate nominal trajectory, the inner planner searches for a feasible maneuver to a designated target. If the inner planner fails due to insufficient samples or difficult geometry, the method cannot distinguish true infeasibility from a search failure. SCRAMPPI retains this trajectory-wise requirement but replaces nested searches with HJ value-function queries.

\subsection{Hamilton-Jacobi Reachability in Planning}

Several methods build on HJ reachability for safe robot navigation. FaSTrack \cite{herbert2017fastrack} bounds tracking error between planning and tracking models, while HJ-Patch \cite{tonkens2024patching} updates safety value functions. DeepReach \cite{bansal2021deepreach} and HJRNO \cite{li2025hjrno} use neural networks to approximate HJ value functions, improving scalability but requiring separate treatment of approximation error. We use GPU-based grid solvers for recovery problems within tractable grid sizes; their numerical values likewise are not exact certificates.

Bajcsy et al. \cite{bajcsy2019efficient} and Herbert et al. \cite{herbert2019reachability} accelerate reachability computations through local updates and warm starts. We adopt Bajcsy et al.'s conservative mapping principle: unknown space is occupied, and observations accumulate confirmed free space. We recompute the finite-horizon reach-avoid problem rather than importing guarantees from their infinite-horizon obstacle-avoidance computations.

\subsection{Safe Sampling-Based MPC}

MPPI \cite{williams2016aggressive,williams2018information} optimizes control sequences by forward-simulating trajectories and computing a cost-weighted average. Several methods incorporate safety via control barrier functions \cite{yin2023shield,gandhi2023safe,rabiee2024guaranteed}. DualGuard MPPI \cite{borquez2025dualguard} uses HJ-based filtering during rollouts and execution for obstacle avoidance. HJ safety and liveness filters \cite{borquez2024safety} also address target-reaching objectives; our focus is evaluating contingency feasibility along nominal rollouts.
Most relevant to our integration strategy is the resampling-based rollout (RBR) method of Yin et al.\ \cite{yin2025safe}. Failed rollouts copy a surviving trajectory's prefix and continue with sampled future controls, avoiding wasted computation. Yin et al.\ use discrete-time neural control barrier functions as the survival criterion. We instead use numerical HJ reach-avoid checks and restrict parent selection to the same proposal group.

\textit{Placement of this work.} Existing contingency planners either target collision avoidance or stopping rather than a designated recovery target~\cite{alsterda2019contingency,tordesillas2022faster}, account for uncertain agent behaviors~\cite{chen2022branch,li2023marc}, or assess recovery feasibility through sampling~\cite{jung2025contingency}. HJ reachability methods support obstacle avoidance~\cite{bajcsy2019efficient,herbert2019reachability,borquez2025dualguard} and safety and liveness filtering~\cite{borquez2024safety}, rather than the trajectory-wise contingency requirement considered here. SCRAMPPI uses finite-horizon HJ value-function queries to evaluate access to a designated recovery target while operating in partially unknown environments. GPU-based solvers make this integration practical within tractable grid sizes. The proposed framework is outlined in Fig.~\ref{fig:process}.

\section{Preliminaries}
\label{sec:preliminaries_new}

\subsection{Finite-Horizon HJ Reach-Avoid Analysis}
\label{sec:hj_background_new}

Consider a system with state $\mathbf{x}\in\mathcal{X}\subseteq\mathbb{R}^{n}$ and dynamics
\begin{equation}
    \dot{\mathbf{x}}=f(\mathbf{x},\mathbf{u},\mathbf{d}),
    \qquad \mathbf{u}\in\mathcal{U}(\mathbf{x}),\quad
    \mathbf{d}\in\mathcal{D}(\mathbf{x}),
    \label{eq:hj_dynamics_new}
\end{equation}
where $\mathbf{u}$ is the control and $\mathbf{d}$ is an adversarial disturbance. The reach-avoid problem is to reach a recovery target $\mathcal{S}=\{\mathbf{x}:\ell(\mathbf{x})\leq0\}$ within a horizon $T$, without entering the avoid set $\mathcal{O}=\{\mathbf{x}:g(\mathbf{x})>0\}$ before arrival. The avoid set contains collision states or other forbidden states, and $\mathcal{S}\cap\mathcal{O}=\varnothing$. Thus $g\leq0$ represents admissible states.

Using backward time $\tau\in[-T,0]$, so that $-\tau$ is the remaining time, define the Hamiltonian
\begin{equation}
    H(\mathbf{x},\mathbf{p})
    =\min_{\mathbf{u}\in\mathcal{U}(\mathbf{x})}
    \max_{\mathbf{d}\in\mathcal{D}(\mathbf{x})}
    \mathbf{p}^{\mathsf T}f(\mathbf{x},\mathbf{u},\mathbf{d}).
    \label{eq:hj_hamiltonian_new}
\end{equation}
Under standard differential-game and regularity assumptions, the robust reach-avoid value $V(\mathbf{x},\tau)$ is the viscosity solution of the double-obstacle variational inequality~\cite{margellos2011hamilton, fisac2015reach,bansal2017hamilton}
\begin{subequations}
\label{eq:hj_vi_new}
\begin{align}
    0=\max\big\{g(\mathbf{x})-V,
    \min\big[&\partial_{\tau}V
    +H(\mathbf{x},\nabla_{\mathbf{x}}V),\nonumber\\
    &\ell(\mathbf{x})-V\big]\big\},
    \label{eq:hj_vi_pde_new}\\
    V(\mathbf{x},0)=\max\{\ell(\mathbf{x}),g(\mathbf{x})\}.
    \label{eq:hj_vi_terminal_new}
\end{align}
\end{subequations}
The condition $V(\mathbf{x},\tau)\leq0$ characterizes states from which a causal control strategy can reach $\mathcal{S}$ within $-\tau$ while avoiding $\mathcal{O}$, despite every admissible disturbance. This is an arrival condition, not a requirement to remain in $\mathcal{S}$ indefinitely. The value depends on $\ell$ and $g$ and is not generally a distance or a recovery time. At differentiability points, an associated feedback control satisfies
\begin{equation}
    \mathbf{u}^{*}(\mathbf{x},\tau)
    \in\arg\min_{\mathbf{u}\in\mathcal{U}(\mathbf{x})}
    \max_{\mathbf{d}\in\mathcal{D}(\mathbf{x})}
    \nabla_{\mathbf{x}}V(\mathbf{x},\tau)^{\mathsf T}
    f(\mathbf{x},\mathbf{u},\mathbf{d}).
    \label{eq:hj_optimal_control_new}
\end{equation}

\subsection{Model Predictive Path Integral Control}
\label{sec:mppi_background_new}

Consider discrete dynamics $\mathbf{x}_{k+1}=F(\mathbf{x}_k,\mathbf{u}_k,\bar{\mathbf{d}}_k)$, where $\bar{\mathbf{d}}_k$ is the disturbance used for prediction, and a length-$N$ control sequence $\mathbf{U}=\mathbf{u}_{0:N-1}$. Model predictive path integral (MPPI) control~\cite{williams2016aggressive,williams2018information} samples $M$ perturbed sequences around a current mean $\bar{\mathbf{U}}$:
\begin{equation}
    \mathbf{U}^{(i)}=\bar{\mathbf{U}}+\boldsymbol{\epsilon}^{(i)},
    \qquad
    \boldsymbol{\epsilon}^{(i)}_k\sim\mathcal{N}(\mathbf{0},\boldsymbol{\Sigma}),
    \quad i=1,\ldots,M,
    \label{eq:mppi_sampling_new}
\end{equation}
where $\boldsymbol{\Sigma}$ is the control-perturbation covariance. These perturbations explore candidate controls; they are not the external disturbance. Each sequence is propagated through $F$ and assigned a trajectory score $J_i$, including task, control, and any sampling-correction terms of the chosen MPPI formulation. Lower-cost samples receive larger weights. With $\rho=\min_i J_i$ and temperature $\lambda>0$, the single-proposal update is
\begin{align}
    w_i&=\frac{\exp\big(-(J_i-\rho)/\lambda\big)}
    {\sum_{j=1}^{M}\exp\big(-(J_j-\rho)/\lambda\big)},
    \label{eq:mppi_weights_new}\\
    \bar{\mathbf{u}}_k^{+}
    &=\bar{\mathbf{u}}_k+\sum_{i=1}^{M}w_i\boldsymbol{\epsilon}^{(i)}_k,
    \qquad k=0,\ldots,N-1.
    \label{eq:mppi_update_new}
\end{align}
Only the first updated control is applied; the sequence is then shifted for the next planning cycle. Hard sampled constraints can be imposed by assigning infinite score to violating rollouts, provided at least one finite-score sample remains. Section~\ref{sec:mppi_new} extends this construction to multiple proposed means and resampling.

\section{Problem Formulation}
\label{sec:problem_formulation_new}

\subsection{Contingency-Constrained Planning}
\label{sec:contingency_problem_new}

The robot uses a nominal model for its task and a contingency model for recovery:
\begin{subequations}
\label{eq:mode_dynamics_new}
\begin{align}
    \dot{\mathbf{x}}&=f_{\mathrm{nom}}(\mathbf{x},\mathbf{u},\mathbf{d}),
    &&\mathbf{u}\in\mathcal{U}_{\mathrm{nom}}(\mathbf{x}),\quad
    \mathbf{d}\in\mathcal{D}_{\mathrm{nom}}(\mathbf{x}),
    \label{eq:nominal_dynamics_new}\\
    \dot{\mathbf{x}}&=f_{\mathrm{con}}(\mathbf{x},\mathbf{u},\mathbf{d}),
    &&\mathbf{u}\in\mathcal{U}_{\mathrm{con}}(\mathbf{x}),\quad
    \mathbf{d}\in\mathcal{D}_{\mathrm{con}}(\mathbf{x}).
    \label{eq:contingency_dynamics_new}
\end{align}
\end{subequations}
These models coincide, as in contingency MPC~\cite{alsterda2019contingency,alsterda2021contingency}, or differ, modeling contingencies that affect the robot's dynamics.

Let $\mathcal{O}_{\mathrm{nom}}$ and $\mathcal{O}_{\mathrm{con}}$ be the nominal and contingency avoid sets. They may include collision states and other forbidden states, such as road departure. The recovery target $\mathcal{S}\subseteq\mathcal{X}\setminus\mathcal{O}_{\mathrm{con}}$ specifies the task's desired contingency outcome and may contain several recovery regions; it need not be defined only by collision avoidance. Over $N$ nominal steps, the objective $J_{\mathrm{nom}}(\mathbf{X},\mathbf{U})$ scores the state sequence $\mathbf{X}=\mathbf{x}_{0:N}$ and controls $\mathbf{U}=\mathbf{u}_{0:N-1}$. The nominal prediction map $F_{\mathrm{nom}}$ propagates $f_{\mathrm{nom}}$ over $\Delta t$ using a specified disturbance $\bar{\mathbf{d}}_k$. This prediction sequence is fixed during optimization, rather than chosen adversarially.

Following Contingency-MPPI~\cite{jung2025contingency}, we require a contingency maneuver from every nominal state, including the current and terminal states:
\begin{subequations}
\label{eq:contingency_problem_new}
\begin{align}
    \min_{\mathbf{U}}\quad
    &J_{\mathrm{nom}}(\mathbf{X},\mathbf{U})
    \label{eq:cost_new}\\
    \mathrm{s.t.}\quad
    &\mathbf{x}_0=\mathbf{x}_{\mathrm{init}},
    \label{eq:initial_state_new}\\
    &\mathbf{x}_{k+1}=F_{\mathrm{nom}}
    (\mathbf{x}_k,\mathbf{u}_k,\bar{\mathbf{d}}_k),
    \label{eq:dynamics_constraint_new}\\
    &\mathbf{u}_k\in\mathcal{U}_{\mathrm{nom}}(\mathbf{x}_k),
    \label{eq:nominal_constraints_new}\\
    &\mathbf{x}_k\notin\mathcal{O}_{\mathrm{nom}},\quad k=0,\ldots,N,
    \label{eq:terminal_constraint_new}\\
    &\forall k\in\{0,\ldots,N\},\ \exists\boldsymbol{\pi}_k
    \label{eq:contingency_exists_new}\\
    &\quad\forall\,\mathbf{d}_k(\cdot)\ \text{admissible},
    \nonumber\\
    &\quad\exists\theta_k\in[0,T_{\mathrm{con}}]:\quad
    \boldsymbol{\xi}_k(\theta_k)\in\mathcal{S},
    \label{eq:contingency_target_new}\\
    &\quad\boldsymbol{\xi}_k(s)\notin\mathcal{O}_{\mathrm{con}},
    \quad\forall s\in[0,\theta_k].
    \label{eq:contingency_admissible_new}
\end{align}
\end{subequations}
Dynamics and input constraints hold for $k=0,\ldots,N-1$. Here $\boldsymbol{\xi}_k$ follows $f_{\mathrm{con}}$ from $\boldsymbol{\xi}_k(0)=\mathbf{x}_k$, under the causal policy $\boldsymbol{\pi}_k$ and disturbance signal $\mathbf{d}_k(\cdot)$. The policy respects $\mathcal{U}_{\mathrm{con}}$, and admissible disturbances take values in $\mathcal{D}_{\mathrm{con}}$ along the trajectory. The last constraints require arrival in $\mathcal{S}$ within $T_{\mathrm{con}}$ while avoiding $\mathcal{O}_{\mathrm{con}}$, for every such disturbance.

This formulation differs from \cite{jung2025contingency} in two major ways: 
1) we consider both nominal and contingency \textit{dynamics}, and 
2) we ensure robustness to disturbances.

\subsection{HJ Representation of Contingency Feasibility}
\label{sec:key_insight_new}

Applying the reach-avoid construction in Section~\ref{sec:hj_background_new} to $f_{\mathrm{con}}$, $\mathcal{U}_{\mathrm{con}}$, $\mathcal{D}_{\mathrm{con}}$, $\mathcal{O}_{\mathrm{con}}$, and $\mathcal{S}$ gives the value function $V_{\mathrm{con}}$. Under the stated assumptions, the contingency constraints are equivalent to membership in the recovery-feasible set
\begin{equation}
\begin{aligned}
    \mathcal{C}_{\mathrm{con}}(T_{\mathrm{con}})
    &:=\{\mathbf{x}:V_{\mathrm{con}}(\mathbf{x},-T_{\mathrm{con}})\leq0\},\\
    \mathbf{x}_k&\in\mathcal{C}_{\mathrm{con}}(T_{\mathrm{con}}),
    \quad k=0,\ldots,N.
\end{aligned}
\label{eq:feasible_set_new}
\end{equation}
Unlike Contingency-MPPI, which relies on an expensive recovery maneuver sampling procedure that scales with both the action space dimension and the time horizon, we avoid this computational burden by replacing sampling with a rapid, single query of the value-function.

\section{Approach}
\label{sec:approach_new}

\subsection{Framework Overview}

SCRAMPPI retains MPPI for nominal planning and replaces Contingency-MPPI's nested recovery searches with HJ value-function queries. The nominal objective ranks candidate trajectories; the queries check whether recovery remains possible from each nominal state. Resampling-based rollouts (RBR) reuse surviving trajectory prefixes to reduce wasted samples. The resulting command is checked before execution, and a task-defined trigger activates the HJ recovery controller. Algorithm~\ref{alg:scramppi} gives the nominal sequence update. The samples are divided approximately evenly among proposal groups $\mathcal G_j$. HJ checks use~\eqref{eq:numerical_feasibility_new} and reject states outside the grid. The returned sequence undergoes command validation in Section~\ref{sec:control_selection_new}; failure is passed to the task-level switching logic.

\begin{algorithm2e}[t]
\DontPrintSemicolon
\small
\KwIn{$\mathbf x_0$, proposal means $\{\boldsymbol\mu^j\}_{j=1}^{G}$, $\widehat V_{\mathrm{con}}$, current map} 
\KwOut{Nominal sequence $\bar{\mathbf U}^{+}$ or failure}

\lIf{$\mathbf x_0$ fails nominal or HJ checks}{\Return failure}
Split $M$ samples into proposal groups $\mathcal G_j$\;
\For{each sample $i$ in group $j$}{
    Draw perturbations $\boldsymbol\epsilon^{(i)}$ as in~\eqref{eq:mppi_sampling_new}\;
    $\widetilde{\mathbf U}^{(i)}\gets\boldsymbol\mu^j+\boldsymbol\epsilon^{(i)},\quad \mathbf x_{i,0}\gets\mathbf x_0$\;
}
\BlankLine
\For{$k=0$ \KwTo $N-1$}{
    \For{each active group $\mathcal G_j$}{
        $\mathbf x_{i,k+1}\gets F_{\mathrm{nom}}(\mathbf x_{i,k},\widetilde{\mathbf u}^{(i)}_k,\bar{\mathbf d}_k),\quad i\in\mathcal G_j$\;
        Check nominal constraints and $\widehat V_{\mathrm{con}}(\mathbf x_{i,k+1},-T_{\mathrm{con}})\leq-\delta$\;
        $\mathcal K\gets$ surviving sample indices in $\mathcal G_j$\;
        \lIf{$\mathcal K=\varnothing$}{discard group; \textbf{continue}}
        \For{each failed sample $i\in\mathcal G_j$}{
            $p\sim\operatorname{Uniform}(\mathcal K)$\;
            $\widetilde{\mathbf u}^{(i)}_{0:k}\gets\widetilde{\mathbf u}^{(p)}_{0:k},\quad \mathbf x_{i,k+1}\gets\mathbf x_{p,k+1}$\;
            Keep the remaining controls of sample $i$\;
        }
    }
}
\BlankLine
Repropagate remaining sequences from $\mathbf x_0$ without resampling; compute costs $J_i$\;
$\mathcal I\gets$ indices passing all nominal and HJ checks\;
\lIf{$\mathcal I=\varnothing$}{\Return failure}
$\rho\gets\min_{i\in\mathcal I}J_i;\quad w_i\gets0\ \text{for }i\notin\mathcal I$\;
$w_i\gets\displaystyle\frac{\exp(-(J_i-\rho)/\lambda)}{\sum_{r\in\mathcal I}\exp(-(J_r-\rho)/\lambda)},\quad i\in\mathcal I$\;
\Return $\bar{\mathbf U}^{+}=\sum_{i\in\mathcal I}w_i\widetilde{\mathbf U}^{(i)}$\;
\caption{SCRAMPPI Algorithm}
\label{alg:scramppi}
\end{algorithm2e}

\subsection{HJ Value-Function Queries}
\label{sec:value_function_new}

Let $\widehat V_{\mathrm{con}}$ be a numerical approximation of the reach-avoid value in~\eqref{eq:feasible_set_new}. A nominal state passes the contingency check when
\begin{equation}
    \widehat V_{\mathrm{con}}(\mathbf{x},-T_{\mathrm{con}})
    \leq-\delta,\qquad \delta\geq0.
    \label{eq:numerical_feasibility_new}
\end{equation}
Each state is tested with the full contingency horizon $T_{\mathrm{con}}$. Queries use the complete contingency state, not only position, and states outside the numerical grid are rejected. The margin $\delta$ allows numerical tightening, but without an approximation-error bound it is not a certificate for the physical system. Time-indexed values are retained for recovery execution.

\textbf{Map updates:} Fixed recovery problems are solved before operation; changes to the dynamics, disturbance bounds, avoid set, or target require an updated solution. For navigation in static environments, we adopt the conservative mapping principle of Bajcsy et al.~\cite{bajcsy2019efficient}: accumulate confirmed free space and treat unobserved space as occupied. With unchanged dynamics, disturbances, and horizon, shrinking the avoid set and retaining or expanding the target cannot remove an existing recovery maneuver. The last completed solution can therefore be used during these updates. This reasoning requires conservative observations that do not revoke previously confirmed free space or targets. If constraints tighten or a target is removed, the old solution no longer justifies continued nominal operation without a new feasibility check.

Unlike the obstacle-avoidance computation in~\cite{bajcsy2019efficient}, we solve a finite-horizon reach-avoid problem. We recompute~\eqref{eq:hj_vi_new} from its terminal condition rather than adopting that work's warm-start or local-update procedure.

\subsection{Multimodal MPPI with Resampling}
\label{sec:mppi_new}

\textbf{Ancillary proposals:} Following Contingency-MPPI~\cite{jung2025contingency}, Topo-PRM generates homotopically distinct workspace paths and nonlinear MPC (NMPC) converts them into dynamically feasible control sequences. An value-function mask excludes workspace cells for which no represented full state passes~\eqref{eq:numerical_feasibility_new}. This guides path generation but does not replace full-state rollout checks. Together with the shifted MPPI mean, the ancillary sequences give $G$ proposal means $\boldsymbol{\mu}^{1},\ldots,\boldsymbol{\mu}^{G}$. The $M$ samples are divided approximately evenly among these groups. A sample in group $j$ is drawn as
\begin{equation}
    \mathbf{U}^{(i)}=\boldsymbol{\mu}^{j}
    +\boldsymbol{\epsilon}^{(i)},
    \qquad i\text{ in group }j,
    \label{eq:multimodal_sampling_new}
\end{equation}
with perturbations as in~\eqref{eq:mppi_sampling_new}. Each sequence is propagated using $F_{\mathrm{nom}}$ and $\bar{\mathbf{d}}_k$. At each step, a sample survives only if its nominal input and state constraints and the value-function check all pass.

\textbf{Group-local RBR:} Simply assigning infinite cost to violating rollouts can leave too few useful samples, especially in high dimensional tight certificates. We use RBR to improve sample efficiency, aiming to reduce the need for larger sample budgets and repeated sampling-covariance tuning without relaxing the feasibility checks. Instead of immediately discarding a failed rollout, Resampling based rollout(RBR)~\cite{yin2025safe} copies the realized control prefix of a randomly selected survivor and continues from that survivor's state, retaining the failed sample's independently drawn suffix. Standard RBR selects parents from the whole population. With multiple proposals, this can replace samples from a difficult route with samples from an easier route. SCRAMPPI restricts parent selection to the same proposal group, preventing one group from repopulating another. If a group has no survivor, its samples remain failed.

After resampling, every resulting sequence $\widetilde{\mathbf U}^{(i)}$ is propagated again from the measured state without rewiring. This checks the complete trajectory and evaluates its nominal cost $J_i$. A sequence that violates any nominal or value-function constraint receives zero weight, not a large finite penalty. Let $\mathcal I$ contain the strictly accepted samples retained for aggregation; an experiment may use all accepted samples or a stated lowest-cost fraction. With $\rho=\min_{i\in\mathcal I}J_i$ and $\mathcal I\neq\varnothing$, the update is
\begin{equation}
\begin{aligned}
    w_i&=\frac{\exp(-(J_i-\rho)/\lambda)}
    {\sum_{j\in\mathcal I}\exp(-(J_j-\rho)/\lambda)},\\
    \bar{\mathbf U}^{+}&=\sum_{i\in\mathcal I}
    w_i\widetilde{\mathbf U}^{(i)}.
\end{aligned}
\label{eq:multimodal_update_new}
\end{equation}
This is a cost-weighted multimodal MPPI update; no proposal-density or resampling-ancestry correction is applied. RBR changes how the rollout budget is used, not the contingency-feasibility condition. After a nominal command is executed, the sequence is shifted for the next cycle.

\subsection{Command Selection and HJ Recovery}
\label{sec:control_selection_new}

Nominal planning first checks the measured state against the nominal constraints and~\eqref{eq:numerical_feasibility_new}. Because averaging accepted sequences need not preserve feasibility, SCRAMPPI also propagates the weighted first control for one step and checks its input, successor state, and contingency feasibility. If this fails, it checks the first control of the lowest-cost accepted sample instead. A failed current-state check, empty accepted set, or failed command check is reported to the task-level switching logic. Such failures activate recovery only if included in the task's trigger rule.

At activation time $t_s$, the recovery controller uses
\begin{equation}
    \tau(t)=t-t_s-T_{\mathrm{con}},
    \qquad t\in[t_s,t_s+T_{\mathrm{con}}],
    \label{eq:recovery_clock_new}
\end{equation}
in~\eqref{eq:hj_optimal_control_new}, with $\widehat V_{\mathrm{con}}$ in place of $V$. Thus the remaining recovery time decreases rather than restarting at each control update. Recovery seeks to reach $\mathcal S$ before this horizon expires; activation and completion are specified with each task.

\section{Simulation Studies}
\label{sec:simulation_studies_new}

\subsection{Experimental Setup}
\label{sec:experiment_protocol_new}

We compare the following methods in both simulated environments:
\begin{itemize}
    \item \textbf{MPPI:} Standard goal-reaching without contingency checks~\cite{williams2016aggressive}.
    \item \textbf{C-MPPI:} Nested MPPI searches for a feasible recovery from each checked nominal state~\cite{jung2025contingency}.
    \item \textbf{CMPC:} Nonlinear MPC jointly optimizes nominal and contingency branches sharing their first input~\cite{alsterda2019contingency}.
    \item \textbf{SCRAMPPI:} Our method replaces nested recovery searches with numerical HJ value-function queries.
\end{itemize}
Within each scenario, MPPI, C-MPPI, and SCRAMPPI use 100 nominal samples and the same sampling covariance. C-MPPI uses 100 inner recovery samples per checked nominal state. All three use RBR and Topo-PRM/NMPC ancillary proposals to match sampling and initialization resources. RBR applies each method's own constraints, leaving MPPI contingency-unaware.

We report the following metrics:
\begin{itemize}
    \item \textbf{\TaskSuccess:} Percentage of episodes reaching the nominal goal within the episode limit without collision or departure from the permitted region.
    \item \textbf{\TTGoal:} Mean time to the nominal goal, averaged over successful episodes.
    \item \textbf{\SafeRate:} Mean number of executed states flagged as contingency-infeasible by the HJ value function per episode; this is a model-based diagnostic.
    \item \textbf{Success$\mid$activated:} Successful task completions among episodes that activate recovery, reported for the ice scenario only.
\end{itemize}

Both scenarios assume zero-disturbance, so these experiments evaluate deterministic numerical contingency feasibility rather than robustness to a nonzero disturbance set. Navigation uses $\delta=0$; the ice experiment uses $\delta=0.30$ as an approximation-error bound.

\subsection{\scenarioOne}
\label{sec:navigation_sim_new}

We consider a unicycle model navigating an unknown environment with states $\mathbf{x}=(X,Y,\psi)$ and controls $(v,\omega)$ that follows
\begin{equation}
    \dot X=v\cos\psi,\qquad
    \dot Y=v\sin\psi,\qquad
    \dot\psi=\omega.
    \label{eq:navigation_dynamics_new}
\end{equation}
We generate 100 cluttered maps and retain only maps containing a collision-free geometric route between the start and goal; this common pre-screen does not test dynamic or contingency feasibility. A limited-range sensor reveals obstacles and recovery regions online in a radius of 20m around the robot. Unknown space is treated as occupied, and confirmed free space and targets are accumulated as described in Section~\ref{sec:value_function_new}. We do not use a specific task or trigger and generate random safe zones to highlight the online HJ value function update as the map updates.

\begin{figure}[t]
    \centering
    \includegraphics[width=0.98\columnwidth]{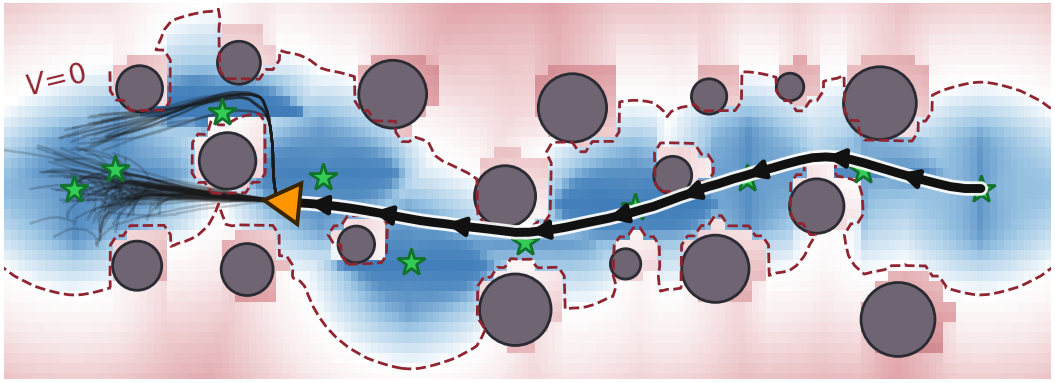}
    \caption{Trajectory and sample rollouts of SCRAMPPI in the \scenarioOne. The heat map shows a workspace projection of the numerical reach-avoid value, while rollout checks use the full state. Green stars denote recovery targets. RBR enables the samples to both be safe and ensure coverage.}
    \label{fig:sim}
\end{figure}

The nominal planning horizon is 30 steps, or 6.0\,s. SCRAMPPI evaluates recovery with $T_{\mathrm{con}}=1.4$\,s on a $200\times80\times40$ grid and nearest-cell queries. C-MPPI uses 100 recovery samples per nominal state. The HJ field is recomputed when the represented obstacle or target set changes. Figure~\ref{fig:sim} shows a representative map, the executed trajectory, and a workspace projection of the value function.

As seen in \Cref{tab:sim_results}, SCRAMPPI completes all 100 tasks and records no states flagged by the common HJ evaluator. C-MPPI reaches 97\% task success but still averages 0.4 unsafe states per episode; MPPI and CMPC attain 88\% and 84\% success. Among contingency-aware methods, SCRAMPPI also has the shortest successful-trial goal time.


\begin{table}[t]
  \centering \scriptsize
  \caption{Simulation results. SCRAMPPI has the highest success rate and lowest safety violations of all methods while being faster than the contingency aware baselines. For Ice Slip, we additionally report the number of successes over the number of episodes our contingency model is activated in, and demonstrate our method recovers in the case of ice slip consistently. }
  \label{tab:sim_results}
  \setlength{\tabcolsep}{7pt}
  \begin{tabular}{@{} l c c c c @{}}
    \toprule
    \multicolumn{5}{c}{\textbf{\scenarioOne \ (100 episodes)}}\\
    \textbf{Metric} & \textbf{MPPI} & \textbf{C-MPPI} & \textbf{CMPC} & \textbf{SCRAMPPI} \\
    \midrule
    \TaskSuccess \ (\%) $\uparrow$ & $88$ & $97$ & $84$ & $\mathbf{100}$ \\
    \TTGoal \ (s) $\downarrow$ & $\mathbf{16.0}$ & $22.4$ & $23.1$ & $20.4$ \\
    \SafeRate \ $\downarrow$ & $15.4$ & $0.4$ & $0.8$ & $\mathbf{0.0}$ \\
    \midrule
    \multicolumn{5}{c}{\textbf{\scenarioTwo \ (25 episodes)}}\\
    \textbf{Metric} & \textbf{MPPI} & \textbf{C-MPPI} & \textbf{CMPC} & \textbf{SCRAMPPI} \\
    \midrule
    \TaskSuccess \ (\%) $\uparrow$ & $12$ & $80$ & $72$ & $\mathbf{100}$ \\
    \TTGoal \ (s) $\downarrow$ & $\mathbf{11.4}$ & $13.7$ & $15.9$ & $12.6$ \\
    \SafeRate \ $\downarrow$ & $18.6$ & $2.4$ & $3.7$ & $\mathbf{0.0}$ \\
    Success $\mid$ activated $\uparrow$ & $3/25$ & $20/24$ & $8/12$ & $\mathbf{23/23}$ \\

    \bottomrule
  \end{tabular}
\end{table}
\subsection{\scenarioTwo}   
\label{sec:ice_experiment_new}

In our ice slip environment, the contingency requirement is to move the vehicle from a roadway with areas of uncertain traction to the shoulder of the road with known traction after loss of control authority is detected. After the known-traction area is reached, nominal planning resumes under the known local model. The environments are generated with randomized patches of ice, obstacles and continuous or alternating shoulders in a 90\degree{} turn.

The lateral-slip planning model is based on  Gao et al.~\cite{gao2010predictive} and augmented with longitudinal acceleration and drag. With state $\mathbf{x}=(X,Y,v_x, v_y,\psi)$ and controls $(a_x,a_y)$, the undisturbed model is
\begin{equation}
\begin{aligned}
    \dot X&=v_x\cos\psi-v_y\sin\psi, &\dot Y&=v_x\sin\psi+v_y\cos\psi, \\
    \dot v_x&=a_x-\kappa(X,Y)v_x, &\dot v_y&=a_y-v_y/\tau_y, \\
    \dot\psi&=a_y/\max(v_x,v_{\mathrm{floor}}),  &a_x^2+a_y^2&\leq a_r(X,Y)^2.
\end{aligned}
\label{eq:ice_dynamics_new}
\end{equation}
where, $\tau_y$ is the lateral-velocity decay time constant, $v_{\mathrm{floor}}>0$ regularizes the yaw-rate expression at low speed, $\kappa(X,Y)$ is the local longitudinal-drag coefficient, $\mu(X,Y)$ is the local friction coefficient. $a_r(X,Y)=\mu(X,Y)g$ is the radius of the local friction-limited acceleration disk which tells us the max acceleration at a position under the local friction. 

We evaluate 25 road and friction configurations. Each planner uses a 40-step, 2.0\,s nominal horizon. SCRAMPPI uses $T_{\mathrm{con}}=2.6$\,s and multilinear queries on a $9\times24\times84\times84\times14$ grid. In addition for SCRAMPPI we assume that, for the computation of the recoverable set, the entire track is covered in ice and only compute this set once offline. The recovery target requires the vehicle footprint to lie inside the known-traction area with bounded $\Delta\psi$, $v_x$, and $v_y$, to ensure recovery. Because of a higher dimensional model, the map is assumed to be known and the value function is computed just once offline.

\begin{figure}[t]
    \centering
    \includegraphics[width=0.98\columnwidth]{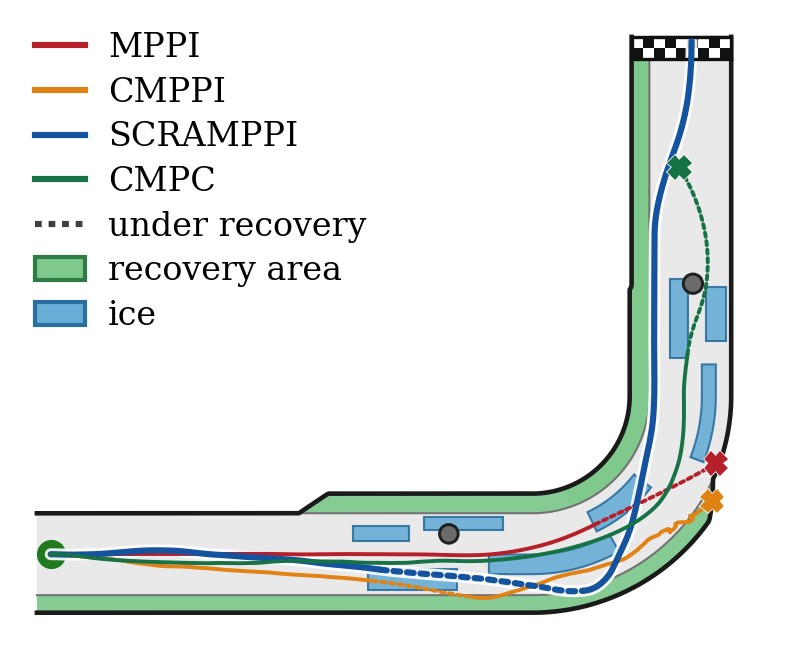}
    \caption{Representative \scenarioTwo \ trial. Solid curves show nominal operation and dotted segments show execution under recovery control. The green band is the known-traction recovery area, blue patches are ice, and gray disks are obstructions. SCRAMPPI is the only method able to navigate the bend while recovering from ice patches, while all other methods either crash or are unable to pull off to the shoulder of the road after slipping.}
    \label{fig:ice_slip}
\end{figure}

The tight bend supports approximately 6.5\,m/s on snow and 4.6\,m/s on ice, while the nominal reference speed is 4.5\,m/s. Low-friction streak locations are hidden from all planners, and recovery activates when the nominal acceleration command lies outside the local friction-limited control disk, i.e., when  $a_x²+a_y² \geq a_r(X,Y)^2$. Because activation depends on the selected commands, the activation denominator differs across methods. Figure~\ref{fig:ice_slip} compares representative trajectories and marks the portions executed under recovery control.

SCRAMPPI completes all 25 tasks and all 23 trials in which recovery activates. C-MPPI completes 80\% of the tasks and 20 of 24 activated trials, while CMPC completes 72\% and 8 of 12. MPPI completes only 12\%. As in navigation, SCRAMPPI records no executed states flagged by the numerical HJ evaluator. 

\subsection{Discussion: Runtime Limitations}
\label{sec:computation_new}


\begin{table}[h]
    \centering
    \small
    \caption{Average Runtime. SCRAMPPI can, on average, run in ~50 ms while C-MPPI takes substantially longer. Of note, we do not need to recompute our HJ value function every planning step, leading to the lower total time}
    \begin{tabular}{p{3.2cm}cc}
        \toprule
         & SCRAMMPI & C-MPPI \\
        \midrule
        HJ Reachability (Min/Avg/Max) (ms) & 18.6 / 55.4 / 181.9 & - \\
        TopoPRM+NMPC (ms) & 41.8 & 57.2 \\
        MPPI Sampling (ms) & 2.6 & 107.7 \\
        Avg Total Time (ms) & 50.2 & 164.9 \\
        \bottomrule
    \end{tabular}
    \label{tab:timing_comparison}
\end{table}

As previous methods \cite{herbert2019reachability} have established, HJ grid-based solvers are generally slow and limited to low-dimensional systems. To allow for online computation, we use the GPU-based \texttt{hj\_reachability} solver~\cite{schmerling_hj_reachability} and batch the value-function queries within MPPI, and run our HJ-computation in a separate thread to our MPPI process. Even for our 5-dimensional ice slip model, we are able to run the HJ computation in 9.69 seconds. In comparison, our 3D model in the \scenarioOne \ takes 0.055 seconds. To compare the speed improvement through computing the reachable set in SCRAMPPI versus nested sampling in C-MPPI, we report the per-component time of each algorithm in the \scenarioOne \ environment in \Cref{tab:timing_comparison}. SCRAMPPI has a significantly faster planning time compared to C-MPPI as it only needs to query the HJ value function (computed once offline). In comparison, C-MPPI takes 40x longer due to increased sampling.

\section{Hardware Experiments}
\label{sec:navigation_hardware_new}

We deploy SCRAMPPI on an AgileX Scout Mini with onboard LiDAR and an NVIDIA RTX~4070 in a $12\times12$\,m workspace containing four obstacles. The robot navigates toward a nominal goal while retaining the ability to hide from a simulated adversary. The adversary's pose is fixed, specified in advance, and known to the robot; no physical adversary is present. Collision-free regions occluded from this viewpoint by confirmed obstacles define the recovery target.

The initial robot pose is chosen so that the nearest obstacle blocks a substantial portion of its LiDAR view, including another obstacle. We use Direct LiDAR--Inertial Odometry (DLIO)~\cite{chen2023dlio}, for LiDAR--inertial localization and mapping. Ray tracing updates the obstacle and occlusion maps as the environment is revealed, with unknown space treated as occupied. The numerical HJ value function is recomputed whenever the represented obstacles or recovery targets change.

The nominal planner uses unicycle dynamics in \eqref{eq:navigation_dynamics_new} with 500 rollouts over 50 steps of $0.2$\,s, giving a $10.0$\,s horizon, with a linear velocity and angular velocity limits of $1.0$\,m/s and $1.0$\,rad/s, respectively. Contingency feasibility is checked at all 50 nominal branch points using $T_{\mathrm{con}}=1.6$\,s, corresponding to eight contingency steps. A separate 100\,Hz low-level feedback controller tracks the planned motion using DLIO odometry.

An operator manually triggers recovery to emulate the adversary becoming alert, at which point the robot must interrupt its nominal task and hide. This is a prescribed task signal, not an autonomous adversary detector. On activation, SCRAMPPI switches to HJ feedback and retreats into an occluded recovery region. Figure~\ref{fig:overview} shows SCRAMPPI choosing a route that retains access to such a region, whereas MPPI would have taken the shorter route without this requirement. 

Across 40 hardware trials, SCRAMPPI reaches the nominal goal in all 20 untriggered trials and reaches an occluded region within $T_{\mathrm{con}}=1.6$\,s of activation in all 20 triggered trials. The complete onboard planning loop averages 15\,Hz, including HJ solve time, supporting real-time operation in this configuration.

\section{Conclusion}
\label{sec:conclusion}

We presented SCRAMPPI, a contingency-constrained planning framework that replaces nested recovery searches with HJ reach-avoid value-function queries while retaining MPPI for the nominal objective. GPU-based solvers and batched queries support real-time nominal planning within tractable grid sizes. Group-local resampling improves use of the rollout budget, while HJ feedback provides a controller for task-defined recovery. Navigation with map revelation, friction-loss recovery, and onboard occlusion-retreat illustrate the intended scope: maintaining access to a recovery target rather than collision avoidance alone.

The main limitation is computational scale. Grid size grows exponentially with state dimension, and larger domains, finer resolution, and longer contingency horizons increase solve costs. The present experiments therefore focus on low-dimensional models and short recovery horizons, which cannot represent distant or slow recovery maneuvers. Numerical approximation error and the cost of updating the value function also limit operation in changing environments.

Learned HJ approximations, including DeepReach~\cite{bansal2021deepreach} and neural operators~\cite{li2025hjrno}, could reduce dependence on dense grids and amortize computation across recovery problems while retaining SCRAMPPI's query-based planning structure. Extending these approximations to richer dynamics, longer horizons, and varying maps could enable aerial landing, energy-aware return maneuvers, and recovery around moving obstacles or agents. Such extensions would require time-varying recovery models where appropriate and validation of value-function and feedback accuracy; neither learned nor grid-based approximations provide an unconditional physical guarantee.

\bibliographystyle{IEEEtran}
\bibliography{references}

\end{document}